\pdfoutput=1
\documentclass[11pt]{article}
\usepackage{amssymb}

\usepackage[preprint]{acl}

\usepackage{listings}

\usepackage{times}
\usepackage{latexsym}
\usepackage{enumitem} % Enables custom numbering

\usepackage{amsmath}
\usepackage{tcolorbox}
\usepackage[T1]{fontenc}
\usepackage[utf8]{inputenc}
\usepackage{titlesec}
\usepackage{url}
\usepackage[stretch=20, shrink=40, expansion=true, spacing=true, tracking=true, protrusion=true]{microtype}
\microtypecontext{spacing=nonfrench}

\usepackage{inconsolata}

\usepackage{graphicx}

\title{MultiMind at SemEval-2025 Task 7:\\ Crosslingual Fact-Checked Claim Retrieval via Multi-Source Alignment}

\author{
\small
Mohammad Mahdi Abootorabi\thanks{Equal contribution.}
\And\small Alireza Ghahramani Kure\footnotemark[1] 
\AND\small Mohammadali Mohammadkhani  \And\small Sina Elahimanesh \And\small Mohammad ali Ali panah \normalfont \small  \AND 
\normalfont
\texttt{\small\{mahdi.abootorabi2, alireza.g.qre,  
mohammadali1379, 
sinaaelahimanesh, 
mohammadali.ap.2000\}@gmail.com} \\
}

\titlespacing*{\paragraph}{0pt}{0.3ex plus 0.1ex minus 0.2ex}{0.5em}
\titleformat{\paragraph}[runin]{\normalfont\normalsize\bfseries}{\theparagraph}{0.4em}{}[]

\begin{document}
\maketitle
\begin{abstract}
This paper presents our system for SemEval-2025 Task 7: Multilingual and Crosslingual Fact-Checked Claim Retrieval. In an era where misinformation spreads rapidly, effective fact-checking is increasingly critical. We introduce \textbf{TriAligner}, a novel approach that leverages a dual-encoder architecture with contrastive learning and incorporates both native and English translations across different modalities. Our method effectively retrieves claims across multiple languages by learning the relative importance of different sources in alignment. To enhance robustness, we employ efficient data preprocessing and augmentation using large language models while incorporating hard negative sampling to improve representation learning. We evaluate our approach on monolingual and crosslingual benchmarks, demonstrating significant improvements in retrieval accuracy and fact-checking performance over baselines.
\end{abstract}

\section{Introduction}
\noindent
The rapid spread of misinformation on online platforms has become a major global challenge \cite{shaar2020known, aimeur2023fake}. False claims can now easily cross linguistic and regional boundaries, magnifying their potential impact \cite{leung2021concerns, fernandez2018online, hakak2021propagation}. The multilingual nature of the Internet amplifies this issue, as misinformation quickly spreads across different language communities, making it harder to track and counteract \cite{bak2023digital}. While professional fact-checkers work tirelessly to verify misleading information, the sheer volume of online content, often published in multiple languages \cite{kazemi2021claim}, makes manual verification increasingly impractical \cite{warren2025show}. The growing speed and scale of misinformation make these efforts more urgent, yet even the most diligent fact-checkers cannot keep pace.

\noindent
Automating the retrieval of fact-checked claims across different languages and cultures is crucial to enhancing the efficiency of fact-checking operations \cite{khurana2023advancing}. The challenge is retrieving relevant fact-checks for social media posts written in languages unfamiliar to fact-checkers \cite{thakur2021beir}. To accurately match claims across linguistic boundaries, systems must recognize claims in different languages and contexts \cite{srba2022fake, dementieva2020evidence} while accounting for subtle language and cultural differences. Given the global nature of misinformation, there is a growing need for multilingual and crosslingual systems that can effectively bridge these gaps \cite{maity2023cross, bontcheva2024evaluating}. By addressing this challenge, we can significantly expedite the process of matching claims to fact-checks, enabling faster and more accurate verification \cite{quelle2023lost}. This challenge motivated our participation in \textit{\textbf{SemEval-2025 Shared Task 7: Multilingual and Crosslingual Fact-Checked Claim Retrieval}} \cite{semeval2025task7}. 

\noindent
In this study, we propose \textbf{TriAligner}, a retrieval pipeline designed to identify relevant fact-checked claims for social media posts across languages. Our system operates in both monolingual and crosslingual settings, reducing the manual effort required for fact-checking. By enhancing multilingual retrieval capabilities, our approach supports fact-checkers, researchers, and media organizations in combating the global spread of misinformation. Our pipeline builds upon the \textit{\textbf{MultiClaim dataset}} \cite{pikuliak-etal-2023-multilingual}, incorporating data augmentation techniques and a dual-encoder architecture designed for the challenges of multilingual claim retrieval. To enhance semantic understanding, we use GPT-4o \cite{hurst2024gpt} to refine post representations, ensuring that fact-checking models can better capture claim-related nuances. Our retrieval system encodes both posts and fact-checks in their native language and English translation, leveraging multilingual embeddings to bridge linguistic gaps. To refine retrieval quality, we introduce a contrastive learning framework that aligns matching claim-post pairs while distinguishing non-matching ones. GPT-4o is further employed as a reranker to refine the relevance of retrieved fact-checks. Our work advances multilingual information retrieval techniques as part of the effort to combat misinformation. We explore how integrating neural retrieval models with contrastive learning and data augmentation can enhance crosslingual fact-checking. Insights from this work are expected to shape the direction of future research in this field. Our team secured the 21st rank in the monolingual setting (Subtask 1) and the 24th rank in the crosslingual setting (Subtask 2) in the test competition.

\begin{figure*}[t]
  \includegraphics[width=\textwidth]{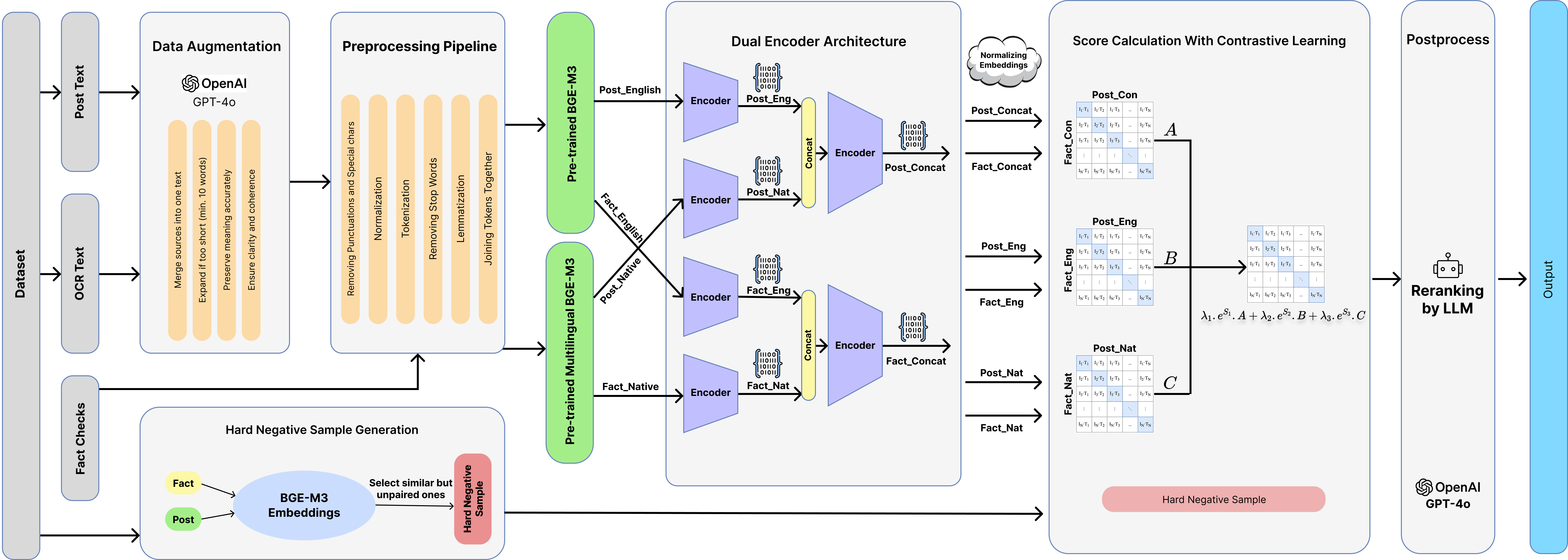}
  \caption{Our proposed system's pipeline includes data augmentation, preprocessing, hard negative sampling, the model's core network, score calculation, and re-ranking using a large language model to generate the final ranking for each social media post.}
  \label{fig:overview}
\end{figure*}

\section{Background and Related Work}
\label{sec:background}
\noindent
The SemEval task features two tracks: monolingual and cross-lingual. In the monolingual track, the post and claim are written in the same language, whereas the cross-lingual track pairs a post with a claim in a different language. 

\noindent
Earlier datasets for previously fact-checked claim retrieval (PFCR), such as CheckThat! \cite{barron2020overview}, primarily focused on English and Arabic, and included manually filtered social media posts to ensure reliability. In contrast, we use the MultiClaim dataset \cite{pikuliak-etal-2023-multilingual}, a large and linguistically diverse resource designed to support multilingual and cross-lingual PFCR. MultiClaim comprises approximately 206k fact-checks in 39 languages and over 28k social media posts in 27 languages. It also includes machine-translated posts and OCR-processed images, further facilitating experimentation across languages and modalities.

\noindent
Regarding methodologies, BM25 and neural text embedding models (TEMs) are commonly used in PFCR tasks. \cite{shaar2020known} relied on BM25 for its efficiency and robustness in monolingual retrieval. Additionally, \cite{sundriyal2023leveraging} applied the integration of neural and traditional approaches to the fact-checking problem on Twitter. However, multilingual and cross-lingual retrieval often requires more sophisticated embedding-based solutions. \cite{pikuliak2023multilingual} shows that embedding models, especially when enhanced with machine translation and supervised fine-tuning, can outperform traditional retrieval methods like BM25 in multilingual contexts. 

\noindent
Recent studies have explored using large language models (LLMs) for multilingual PFCR. \cite{vykopal2025large} evaluated seven LLMs across 20 languages, finding that while LLMs perform well for high-resource languages, they struggle with low-resource ones. Translating texts into English notably improved performance for low-resource languages. Similarly, \cite{singhal2024multilingual} assessed the multilingual fact-checking abilities of five LLMs across five languages using various prompting techniques. They found that zero-shot prompting with self-consistency decoding was most effective, and interestingly, LLMs showed better fact-checking performance in low-resource languages, suggesting potential for mitigating language disparities in PFCR tasks. Additional research has extended PFCR by incorporating modalities such as visual data \cite{mansour2022did}, abstractive summarization \cite{bhatnagar2022harnessing}, and key sentence identification \cite{sheng2021article} to improve retrieval accuracy and contextual understanding.

\section{System Overview}
\label{sec:system}
\noindent
Figure \ref{fig:overview} provides an overview of our method. We employ various techniques to enhance model performance, which we detail in the following sections:
\subsection{Preprocessing Data}
\paragraph{Data Augmentation Using a Large Language Model}
We leverage GPT-4o \cite{hurst2024gpt} large language model (LLM) to augment and rewrite posts by integrating text and OCR data into a cohesive version. Using a predefined prompt (Appendix ~(\S\ref{appendix:prompt})), we enhance each post to improve the model’s ability to comprehend content. After augmentation, each post contains at least 10 words, with all sources merged into a unified text while preserving the original meaning. The augmented and cleaned dataset is publicly available on Hugging Face \footnote{\url{https://huggingface.co/datasets/MultiMind-SemEval2025/Augmented_MultiClaim_FactCheck_Retrieval}}.

\paragraph{Data Cleaning and Preprocessing}
We apply multiple preprocessing steps that are usually used in NLP approaches to improve the model's ability to interpret data effectively. One key step involves concatenating the title and OCR fields for posts and the title and claim fields for facts, creating unified text representations.

\paragraph{Hard Negative Sampling}
To enhance model robustness, we introduce hard negative samples. Using the BGE-M3 \cite{chen-etal-2024-m3}, we encode all facts and posts and identify unrelated pairs with similar representations. These hard negatives help the model better distinguish between semantically similar but unrelated claims and posts.

\paragraph{Generating Embeddings for Facts and Posts}
\label{sec:embeddinggeneration}
Since posts and facts exist in both English and their native languages, we encode them using both English-specific and multilingual pretrained encoders. This process produces four types of embeddings: \textbf{(i)} fact\_native (fact-checked claim in its original language), \textbf{(ii)} fact\_english (fact-checked claim translated to English),
\textbf{(iii)} post\_native (post in its original language),
\textbf{(iv)} post\_english (post translated to English).

\subsection{Core Network of TriAligner}
\noindent
We employ a dual-encoder architecture to independently encode posts and fact-checked claims, drawing inspiration from CLIP \cite{radford2021learning} to align representations across different modalities and spaces.

\noindent
We first utilize pretrained embeddings and map them into a shared semantic space using a neural network encoder in a lower-dimensional space. Next, we concatenate representations from both English and native sources separately, forming unified representations for each post and fact-checked claim. These concatenated encoded vectors then pass through an additional neural encoder to generate compact representations suitable for similarity computation. 

\noindent
For final scoring, after normalization, we compute cosine similarity between embeddings to construct three similarity matrices: \textbf{(i)} Native post and fact-checked claim embeddings, \textbf{(ii)} English post and fact-checked claim embeddings, and \textbf{(iii)} Concatenated embeddings processed through an additional encoder. These correspond to three similarity matrices: \textbf{(A)} Concatenated posts and fact-checked claims, \textbf{(B)} English posts and fact-checked claims, and \textbf{(C)} Native posts and fact-checked claims.

\noindent
To compute the final similarity matrix, we apply Formula \ref{eqn:eq1}, where $x_{i, j}$ represents the element in the $i$th row and $j$th column of the final matrix. Here, $\lambda_1, \lambda_2, \lambda_3$ are trainable coefficients that adjust the weights between different sources, and $S_1, S_2, S_3$ are scaling coefficients for each matrix. The value $x_{i, j}$ corresponds to the similarity score between the $i$th fact-checked claim and the $j$th post. In this way, we use three sources to compute the final similarity score between fact-checked claims and posts.
\vspace{-0.3em}
\begin{equation}
\label{eqn:eq1}
\small
    x_{i, j} = \lambda_1.e^{S_1}.(A_{i, j}) + \lambda_2.e^{S_2}.(B_{i, j}) + \lambda_3.e^{S_3}.(C_{i, j})
\end{equation}
\vspace{-1.5em}

\noindent
We train the model using a symmetric contrastive loss applied to the final similarity matrix \(X \in \mathbb{R}^{N \times N}\), where \(N\) is the batch size. 
This loss encourages true post–claim pairs (diagonal elements $x_{ii}$) to have higher similarity scores while discouraging mismatched pairs (off-diagonal elements 
% \( x_{ij},\ i \neq j \)). 
$x_{ij}, i \neq j$).
Given the element \(x_{ij}\) derived from Equation~\ref{eqn:eq1}, we compute row-wise and column-wise softmax probabilities:
\begin{equation}
\small
P_{ij} = \frac{\exp(x_{ij})}{\sum_{k=1}^N \exp(x_{ik})}, \quad Q_{ij} = \frac{\exp(x_{ij})}{\sum_{k=1}^N \exp(x_{kj})}
\end{equation}
% \begin{small}
% \[
% P_{ij} = \frac{\exp(x_{ij})}{\sum_{k=1}^N \exp(x_{ik})}, \quad
% Q_{ij} = \frac{\exp(x_{ij})}{\sum_{k=1}^N \exp(x_{kj})}
% \]
% \end{small}
The loss is then formulated as the average negative log probability of the true pairs across both perspectives:
\vspace{-0.4em}
\begin{equation}
\label{eqn:eq2}
\small
% \mathcal{L}_{\mathrm{CLIP}}
\mathcal{L}
= -\frac{1}{2N}\sum_{i=1}^N \bigl(\log P_{ii} + \log Q_{ii}\bigr)
\end{equation}
\noindent
This training process and objective preserve residual information, enhance robustness, and enable the model to learn optimal weightings across embedding sources while effectively distinguishing between relevant and irrelevant pairings based on multi-source similarity signals.

\paragraph{Candidate Re-ranking} In the final stage of our pipeline, we employ the GPT-4o model \cite{hurst2024gpt} as a reranker to refine the ordering of candidate fact-checks. We created and utilized a tailored prompt (Appendix ~(\S\ref{appendix:promptreranking})), guiding GPT-4o to rank these candidates strictly based on their relevance to the content of the associated post, producing an output list sorted in descending order of relevance. For each post, we provide the model with the post content and 15 pre-selected candidate fact-checks, instructing it to return the 10 most relevant ones. Since the reranker operates on a fixed pool of 15 candidates, metrics such as Success@20 and Recall@20 remain unchanged with or without the reranking step.

\section{Experimental Evaluation and Results}
\label{sec:experiment}
\noindent
For evaluation, we perform the retrieval task based on the alignment scores between query social media posts and fact-checked claims. The system returns the top-K most relevant fact-checked claims for each social media post in both monolingual and crosslingual configurations. In the monolingual setting, the model retrieves relevant fact-checked claims from the same language as the query post, while in the crosslingual setting, the model searches across fact-checked claims in all available languages.

\noindent
For the architecture of our core TriAligner system, the first layer consists of four encoders with similar structures, each processing different language sources for posts and facts. Each encoder includes a linear layer that reduces the input dimension from 1024 to 256, followed by batch normalization, a ReLU activation function, dropout with a probability of 0.2, and a final linear layer that preserves the 256-dimensional representation. In the second stage, after concatenating the outputs from the previous layer's native and English translation sources, the processed embeddings are fed into additional encoders. Each of these encoders contains a linear layer that reduces dimensions from 512 to 256, followed by a ReLU activation function and another linear layer that maintains the 256-dimensional representation.
Finally, the outputs of the post encoders and initial encoders are normalized, and all are used in the matrix construction phase, where the relevance scores between facts and posts are computed. Due to the dataset size, we did not make the encoders too complex to avoid overfitting. Computational constraints required us to construct the final similarity matrix incrementally by processing the data in batches. For more details of our experiment setup and hyperparameters, see Appendix ~(\S\ref{appendix:experiment}).

\noindent
We evaluated our model with K values of 1, 10, and 20, reporting two metrics: Success@K (S@K) and Recall@K (R@K). Success@K equals 1 when at least one associated fact-check for a post is retrieved in the top K results, and 0 otherwise. Recall@K is more stringent, measuring the proportion of relevant fact-checks retrieved in the top K. For example, if a social media post has two associated fact-checks and only one appears in the top K results, the Recall@K score for that post is 0.5. The formal definitions of these metrics are as follows:
\vspace{-0.4em}
\begin{align}
\scriptstyle 
\text{S@K} = \frac{\text{\# queries with at least one relevant item in top K}}{\text{\# queries}}
\end{align}
\vspace{-1.2em}
\begin{align}
\scriptstyle
\text{R@K} = \frac{1}{\text{\# queries}} \sum_{q=1}^{\text{\# queries}} \frac{\text{\# items in top K that are relevant to query q}}{\text{\# relevant items for query q}}
\end{align}

% To identify the top K elements for each social media post, we constructed a similarity score matrix between all social media posts and fact-checks. In this matrix, each row represents a social media post, and entry $(i,j)$ corresponds to the similarity score between social media post $i$ and fact-check $j$. 

\noindent
Table \ref{table_results} shows the average performance of different model variants in both monolingual and crosslingual settings, without language-specific breakdowns. Tables \ref{table_monolingual} and \ref{table_crosslingual} in Appendix (\S\ref{appendix:tables}) provide detailed language-specific results for monolingual and crosslingual evaluations, respectively, across different stages of our approach. Table \ref{tab:winner_comparison_mono} in Appendix (\S\ref{appendix:tables}) also compares our model's performance against the top-performing team in the test competition.
The evaluation stages are as follows:

\subsection{Baselines Evaluation}
\noindent
For the baselines, we use two powerful pretrained encoders capable of producing informative semantic embeddings at the sentence level:

\paragraph{BGE-M3 \cite{chen-etal-2024-m3}} A versatile embedding model trained through self-knowledge distillation. This method integrates relevance scores from different retrieval functionalities as teacher signals to enhance training quality. The model was trained on multiple datasets covering over 100 languages, making it highly effective for multilingual tasks. The English version also demonstrates excellent performance. BGE-M3 supports dense retrieval, multi-vector retrieval, and sparse retrieval within a unified framework. It can process inputs of varying granularities, from short sentences to long documents of up to 8192 tokens, making it appropriate for our social media posts and fact-checks. It has demonstrated state-of-the-art performance on various benchmarks and tasks.

\begin{table*}[t!]
    \centering
    \caption{Evaluation results across stages for monolingual and crosslingual settings on the development dataset.}
    \resizebox{\textwidth}{!}{%
    \begin{tabular}{c|cccccc|cccccc}
        \multicolumn{13}{c}{Stage 1: Baseline Evaluation Results} \\
        \hline
        & \multicolumn{6}{c|}{Monolingual} & \multicolumn{6}{c}{Crosslingual} \\
        Model/Source of Data & R@1 & R@10 & R@20 & S@1 & S@10 & S@20 & R@1 & R@10 & R@20 & S@1 & S@10 & S@20 \\
        \hline
        LaBSE/Native & 0.276 & 0.576 & 0.625 & 0.303 & 0.598 & 0.645 & 0.074 & 0.255 & 0.286 & 0.082 & 0.274 & 0.304 \\
        LaBSE/English & 0.262 & 0.552 & 0.608 & 0.286 & 0.575 & 0.628 & 0.111 & 0.336 & 0.400 & 0.129 & 0.366 & 0.420 \\
        BGE-M3/English & 0.411 & 0.776 & 0.819 & 0.446 & 0.794 & 0.835 & 0.195 & 0.473 & 0.549 & 0.219 & 0.495 & 0.565 \\
        BGE-M3/Native & 0.410 & \textbf{0.794} & \textbf{0.836} & 0.444 & 0.808 & \textbf{0.848} & 0.142 & 0.392 & 0.434 & 0.158 & 0.409 & 0.448 \\
        BGE-M3/Data Augmentation & \textbf{0.426} & 0.792 & 0.832 & \textbf{0.462} & \textbf{0.811} & 0.847 & \textbf{0.214} & \textbf{0.533} & \textbf{0.603} & \textbf{0.241} & \textbf{0.554} & \textbf{0.616} \\
    \end{tabular}
    }
    
    \vspace{1em}
    \resizebox{\textwidth}{!}{%
        \begin{tabular}{c|cccccc|cccccc}
            \multicolumn{13}{c}{Stage 2: Ablation Study Results} \\
            \hline
            & \multicolumn{6}{c|}{Monolingual} & \multicolumn{6}{c}{Crosslingual} \\
            Idea & R@1 & R@10 & R@20 & S@1 & S@10 & S@20 & R@1 & R@10 & R@20 & S@1 & S@10 & S@20 \\
            \hline
            ConcatEnc & \textbf{0.486} & \textbf{0.816} & \textbf{0.855} & \textbf{0.529} & \textbf{0.827} & \textbf{0.863} & \textbf{0.391} & \textbf{0.680} & \textbf{0.713} & \textbf{0.424} & \textbf{0.694} & \textbf{0.726} \\
            MultiSim & 0.380 & 0.741 & 0.790 & 0.418 & 0.756 & 0.803 & 0.321 & 0.651 & 0.700 & 0.357 & 0.665 & 0.710 \\
        \end{tabular}%
    }

    \vspace{1em}
    \resizebox{\textwidth}{!}{%
        \begin{tabular}{c|cccccc|cccccc}
            \multicolumn{13}{c}{Stage 3: Final System Evaluation Results} \\
            \hline
            & \multicolumn{6}{c|}{Monolingual} & \multicolumn{6}{c}{Crosslingual} \\
            Model & R@1 & R@10 & R@20 & S@1 & S@10 & S@20 & R@1 & R@10 & R@20 & S@1 & S@10 & S@20 \\
            \hline
            TriAligner & 0.501 & 0.837 & 0.879 & 0.545 & 0.848 & 0.886 & 0.367 & 0.687 & 0.720 & 0.402 & 0.707 & 0.728 \\
            TriAligner + Data Augmentation & 0.502 & 0.860 & 0.885 & 0.545 & 0.871 & 0.893 & 0.368 & 0.702 & 0.759 & 0.400 & 0.719 & 0.772 \\
            TriAligner + Augmentation + Re-Ranker & \textbf{0.541} & \textbf{0.870} & \textbf{0.885} & \textbf{0.585} & \textbf{0.881} & \textbf{0.893} & \textbf{0.391} & \textbf{0.734} & \textbf{0.759} & \textbf{0.429} & \textbf{0.748} & \textbf{0.772} \\
        \end{tabular}%
    }

    \label{table_results}
\end{table*}

\paragraph{LaBSE \cite{feng-etal-2022-language}} A multilingual BERT-based sentence embedding model trained on 17 billion monolingual sentences and 6 billion bilingual sentence pairs. The model combines masked language modeling (MLM) and translation language modeling (TLM) pre-training on a 12-layer transformer with a 500K token vocabulary, followed by fine-tuning on a translation ranking task. LaBSE covers over 109 languages within a shared embedding space, enabling crosslingual tasks like semantic search. It shows strong performance even for low-resource languages with limited training data.

\noindent
We leverage these models without additional training by simply passing all posts and fact-checks through them to obtain embeddings. We then perform similarity ranking using cosine similarity to identify the most relevant fact-checks for posts. As demonstrated in Table \ref{table_results}, BGE-M3 outperforms LaBSE, providing superior representations for our task. Our results show that using content in its native language is somewhat more effective in monolingual settings, while in crosslingual settings, English translations perform slightly better. This suggests that multilingual encoders still have room for improvement in cross-language alignment, and translation-to-English pipelines remain more effective despite potential information loss. As Table \ref{table_crosslingual} suggests, the performance gap between English and native language processing is substantial for low-resource languages like Thai, while for well-represented languages, this gap is negligible or even favors native processing. Finally, our evaluation of the refined dataset using BGE-M3 outperformed other baselines, demonstrating the effectiveness of our LLM-assisted refinement method.

\subsection{Ablation Study}
\noindent
To assess the impact of different components in our system, we conducted ablation experiments focusing on two key approaches:

\paragraph{Concatenation-Based Encoding (ConcatEnc)} 
This experiment evaluates the effectiveness of using only the similarity matrix derived from the concatenated embeddings. Instead of using all three similarity matrices as in our full system, we rely solely on the matrix generated from the concatenated native and English representations after they pass through the additional encoder. This allows us to isolate the contribution of the concatenation component to the overall performance without the influence of the individual language-specific matrices.

\paragraph{Multiple Similarity Scoring (MultiSim)} 
This approach assesses the effectiveness of our scoring mechanism in aligning native and English dimensions without using concatenation. MultiSim focuses solely on combining similarity matrices from different language sources. The model produces two separate similarity matrices: one for native-language embeddings and another for English translations. The final similarity matrix is then computed as a linear combination of these matrices using trainable coefficients.

\noindent
Both approaches demonstrate performance improvements over the baselines. The Concatenation-Based Encoding (ConcatEnc) approach yields superior results, as it introduces additional parameters that enable the model to learn more complex dependencies between posts and fact-checks. The fusion of different language representations through concatenation proves to be an effective mechanism for improving retrieval.
More importantly, both methods significantly enhance performance in crosslingual settings by leveraging both native and translated sources. This suggests that aligning multilingual representations contributes to better retrieval capabilities. Given the observed improvements, we incorporate both techniques into our final model.

\subsection{Final System Evaluation}
\noindent
In this stage, we integrate both proposed ideas to develop our final model, TriAligner. As described in Section \ref{sec:system}, the final similarity matrix is a weighted combination of 3 sources: English, native, and fused shared space representations. We train the model with our augmented dataset and subsequently apply an LLM-based re-ranker to enhance performance.
Moreover, we employ hard negative sampling to enhance training and incorporate marginal loss and contrastive loss. However, due to the large batch size, hard negative sampling did not yield significant performance improvements.

\noindent
As shown in Table \ref{table_results}, combining both techniques results in the best performance across all metrics in both monolingual and crosslingual settings, demonstrating the impact of our alignment strategies. Furthermore, LLM-assisted data augmentation provides an additional performance boost, particularly in S@10 and S@20, indicating more effective handling of outlier posts with limited descriptive information. The LLM-based re-ranker also improves performance in both multilingual and crosslingual settings by approximately 5 to 10 percent across different metrics. We anticipate that employing LLMs specifically tailored for advanced reasoning, such as Claude 3.7 Sonnet Thinking \cite{anthropic_claude_3_7_sonnet_2025} or Gemini 2.5 Pro \cite{google_gemini_2_5_pro_2025}, could lead to even greater performance gains. However, due to the limitations of our available computational resources, we employed GPT4-o \cite{hurst2024gpt} for this purpose. 

\noindent
Tables \ref{table_monolingual} and \ref{table_crosslingual} provide detailed language-specific results that offer deeper insights into our system's performance. Languages such as Arabic, Malay, and French benefited significantly from data augmentation. Conversely, augmentation appeared detrimental for the German language, leading to decreased performance across most metrics. This suggests that while the augmentation technique often helps by enhancing sparse or unclear posts, its effectiveness can be language-dependent, potentially due to the LLM's handling of specific linguistic nuances or the nature of the original data for that language. 
Beyond data augmentation, the addition of the LLM-based re-ranker further improved retrieval performance across most languages in both monolingual and crosslingual settings. Notably, the re‑ranker yields substantial gains in Recall@1 and Score@1, underscoring its ability to surface the single most relevant fact‑check by leveraging deeper semantic understanding.
However, the Malay language exhibited a notable decrease in performance with re-ranking. This divergence may stem from frequent code-mixing between Malay and English in social media posts, creating translation inconsistencies during cross-lingual re-ranking, or from the re-ranker's struggle with Malay's narrative-style debunking patterns that differ from Western fact-check templates.
Both tables' baseline results (Stage 1) also highlight that while native embeddings offer slight advantages in some settings, English translations are crucial, particularly for low-resource languages. For instance, the large crosslingual performance gap for the Thai language highlights persistent challenges in multilingual encoder alignment and the continued utility of translation pipelines.

\section{Conclusion}
\label{sec:conclusion}
\noindent
In this work, we introduce TriAligner, a contrastive learning-based approach for multilingual and crosslingual fact-checking, leveraging a dual encoder setup and hard negative samples to improve fact-checked claim retrieval. Through data preprocessing and augmentation, our method improves robustness across diverse languages and social media contexts. Experimental results demonstrate the effectiveness of our approach in retrieving relevant evidence and mitigating misinformation. Future work can explore integrating additional modalities, refining negative sampling strategies, and adapting the model to evolving misinformation patterns. 
Our findings highlight the value of contrastive learning for fast and accurate fact-checking in a globally connected digital landscape. A detailed discussion of limitations and further potential future directions can be found in Appendix ~(\S\ref{appendix:limitation}).

\bibliography{acl_latex}

\newpage
\appendix

\section{Prompt for Data Augmentation}
\label{appendix:prompt}
In this section, we provide the prompt we used for data augmentation:

\begin{tcolorbox}[colback=gray!10, colframe=black, fontupper=\small]
You are provided with 10 pairs of texts, each originating from the same social media post. For each pair, your task is to integrate the two sources into a single cohesive and enhanced text that best represents the content of the post. Combine the information from both the image and the text, rewrite the content to be meaningful, and preserve the post's original context and intent. 

Rules:
\begin{itemize}
    \item You should process pairs individually, ensuring each is handled independently of the others.
    \item The output should be in the language of the post and in the same narrative style.
    \item Do not use phrases like 'The post indicates...'.
    \item Convert abbreviations to their complete form.
    \item Remove hashtags and their tags.
    \item There should not be anything enclosed in brackets, such as [USER] or [URL].
    \item If the combined content is less than ten words, expand it to at least 15 words while staying relevant.
\end{itemize}
\end{tcolorbox}

\newpage

\section{Prompt for Reranking}
\label{appendix:promptreranking}

In this section, we present the prompt we used to rerank the retrieved results with GPT-4o:

\begin{tcolorbox}[colback=gray!10, colframe=black, fontupper=\small]
You are assisting with a fact-check retrieval system that uses neural networks to retrieve relevant fact-checks for social media posts. The current retrieval system returns 20 candidate fact-checks for each post, but its ranking is not perfect. Your task is to re-rank these candidate fact-checks for a single social media post so that the most relevant ones appear at the top.

\#\#\# Task:
- Re-rank the candidate fact-checks based on their relevance to the post’s content.
- Select the top 10 fact-checks from the 15 provided.
- Order the fact-check IDs in descending order of relevance (i.e., the most relevant fact-check appears first).
- Output only the fact-check IDs.

\#\#\# Input Format:
You will receive a dictionary representing a single social media post along with its candidate fact-checks. The structure is as follows:

\begin{lstlisting}
sample_input = {
  "post": {
      "post_id": post_id, 
      "post_content": post_content
  },
  
  "factChecks": [
    {"fact_id": fact_id_1, 
    "fact_content": fact_content_1},
    {"fact_id": fact_id_2, 
    "fact_content": fact_content_2},
    {"fact_id": fact_id_3, 
    "fact_content": fact_content_3},
    ...
  ]
}
\end{lstlisting}

\#\#\# Output Format:
Return a JSON object with a single key (the post\_id) and its value as a list of the top-10 re-ranked fact-check IDs. For example:

\begin{lstlisting}
sample_output = {
  "post_id": [
    fact_id_5,
    fact_id_2,
    fact_id_9,
    ... (total 10 items)
  ]
}
\end{lstlisting}

\#\#\# Important:
- The output must include only fact-check IDs, with no additional scoring information.
- The list must contain exactly 10 fact-check IDs, sorted in descending order of relevance.
- Follow this format strictly.
\end{tcolorbox}

\section{Limitations and Future Direction}
\label{appendix:limitation}
\noindent
In this work, we aimed to address the challenge of fact-check retrieval for social media posts. While our model demonstrates strong performance, certain limitations persist.

First, our crosslingual pipeline can be further improved. Future research should explore more advanced and effective architectures to bridge the gap between different languages, ensuring better crosslingual alignment in a shared representation space.

Second, our approach relies on only two backbone models (BGE-M3 and LaBSE). This limitation can be addressed in future studies by experimenting with a wider range of backbone models to enhance robustness and generalizability.

Lastly, due to resource constraints, we were only able to apply data augmentation to English posts. Future work can extend this augmentation process to other sources (facts) and other languages, further improving the model’s performance across diverse linguistic contexts.

\section{Experiment Setup and Hyperparameters}
\label{appendix:experiment}
\noindent
All experiments were conducted on a single NVIDIA P100 GPU with a batch size of 10000. Our implementation leverages the PyTorch Lightning and Transformers libraries. We trained all model variants using the AdamW optimizer \citep{loshchilov2019decoupled} with a learning rate of $6e-4$ and a cosine annealing learning rate scheduler \cite{loshchilov2017sgdrstochasticgradientdescent}. The initial scaling factor is set to $log(1 / 0.07)$. The training was terminated using early stopping with patience of 5 epochs, monitoring the Recall@10 on the validation set.

\section{Supplementary Evaluation Results}
\label{appendix:tables}

\noindent
In this section, we present evaluation results at different stages for both crosslingual (Table \ref{table_crosslingual}) and monolingual (Table \ref{table_monolingual}) settings for the development dataset. We focus on the eight most frequently used languages in the dataset, as the data for other languages was insufficient for meaningful analysis.

\noindent
Table \ref{tab:winner_comparison_mono} presents a performance comparison between our model and the first-place team in the competition's monolingual setting. It is important to note that our model was evaluated without the re-ranker module during the competition's test phase due to computational resource limitations. We observe that in the separate crosslingual subtask of the test phase of the competition, our system achieved an S@10 score of 0.489, whereas the winning team attained a score of 0.859.

\begin{table*}[t!]
    \centering
    \caption{Evaluation results across different stages for monolingual setting, covering the eight most frequently used languages in the dataset.}
    \resizebox{\textwidth}{!}{%
        \begin{tabular}{c|cccccc|cccccc|cccccc|cccccc}
            \multicolumn{25}{c}{Stage 1: Baseline Evaluation Results} \\
            \hline
            & \multicolumn{6}{c|}{Arabic (ara)} & \multicolumn{6}{c|}{German (deu)} & \multicolumn{6}{c|}{English (eng)} & \multicolumn{6}{c}{French (fra)} \\
            Model/Source of Data & R@1 & R@10 & R@20 & S@1 & S@10 & S@20 & R@1 & R@10 & R@20 & S@1 & S@10 & S@20 & R@1 & R@10 & R@20 & S@1 & S@10 & S@20 & R@1 & R@10 & R@20 & S@1 & S@10 & S@20 \\
            \hline
            LaBSE/Native & 0.346 & 0.692 & 0.756 & 0.346 & 0.692 & 0.756 & 0.193 & 0.440 & 0.488 & 0.241 & 0.470 & 0.518 & 0.162 & 0.435 & 0.478 & 0.193 & 0.464 & 0.508 & 0.498 & 0.711 & 0.727 & 0.516 & 0.713 & 0.729 \\
            BGE-M3/English & 0.436 & 0.795 & 0.859 & 0.436 & 0.795 & 0.859 & 0.331 & 0.711 & 0.777 & 0.398 & 0.747 & 0.807 & 0.309 & 0.737 & 0.782 & 0.349 & 0.759 & 0.801 & 0.678 & 0.887 & 0.903 & 0.702 & 0.888 & 0.904 \\
            BGE-M3/Native & 0.487 & 0.846 & 0.859 & 0.487 & 0.846 & 0.859 & 0.241 & 0.663 & 0.741 & 0.301 & 0.687 & 0.771 & 0.257 & 0.705 & 0.751 & 0.295 & 0.728 & 0.772 & 0.657 & 0.844 & 0.876 & 0.681 & 0.846 & 0.878 \\
            BGE-M3/Data Augmentation & 0.359 & 0.769 & 0.821 & 0.359 & 0.769 & 0.821 & 0.307 & 0.681 & 0.747 & 0.386 & 0.723 & 0.783 & 0.313 & 0.733 & 0.782 & 0.352 & 0.762 & 0.803 & 0.676 & 0.879 & 0.895 & 0.697 & 0.883 & 0.899 \\
            \hline
            & \multicolumn{6}{c|}{Malay (msa)} & \multicolumn{6}{c|}{Portuguese (por)} & \multicolumn{6}{c|}{Spanish (spa)} & \multicolumn{6}{c}{Thai (tha)} \\
            Model & R@1 & R@10 & R@20 & S@1 & S@10 & S@20 & R@1 & R@10 & R@20 & S@1 & S@10 & S@20 & R@1 & R@10 & R@20 & S@1 & S@10 & S@20 & R@1 & R@10 & R@20 & S@1 & S@10 & S@20 \\
            \hline
            LaBSE/Native & 0.167 & 0.586 & 0.667 & 0.181 & 0.600 & 0.676 & 0.237 & 0.651 & 0.695 & 0.275 & 0.689 & 0.725 & 0.336 & 0.618 & 0.675 & 0.361 & 0.636 & 0.691 & 0.310 & 0.476 & 0.524 & 0.310 & 0.476 & 0.524 \\
            BGE-M3/English & 0.318 & 0.821 & 0.862 & 0.352 & 0.829 & 0.867 & 0.257 & 0.731 & 0.766 & 0.301 & 0.768 & 0.798 & 0.478 & 0.785 & 0.835 & 0.509 & 0.797 & 0.846 & 0.857 & 0.905 & 0.929 & 0.857 & 0.905 & 0.929 \\
            BGE-M3/Native & 0.332 & 0.859 & 0.906 & 0.352 & 0.867 & 0.914 & 0.290 & 0.795 & 0.843 & 0.331 & 0.818 & 0.858 & 0.526 & 0.844 & 0.876 & 0.561 & 0.852 & 0.886 & 0.619 & 0.833 & 0.929 & 0.619 & 0.833 & 0.929 \\
            BGE-M3/Data Augmentation & 0.405 & 0.835 & 0.848 & 0.438 & 0.848 & 0.857 & 0.297 & 0.770 & 0.811 & 0.348 & 0.805 & 0.834 & 0.503 & 0.823 & 0.863 & 0.538 & 0.833 & 0.873 & 0.786 & 0.952 & 0.952 & 0.786 & 0.952 & 0.952 \\
            \hline
        \end{tabular}%
    }
    
    \vspace{1em}
    \resizebox{\textwidth}{!}{%
        \begin{tabular}{c|cccccc|cccccc|cccccc|cccccc}
            \multicolumn{25}{c}{Stage 2: Ablation Study Results} \\
            \hline
            & \multicolumn{6}{c|}{Arabic (ara)} & \multicolumn{6}{c|}{German (deu)} & \multicolumn{6}{c|}{English (eng)} & \multicolumn{6}{c}{French (fra)} \\
            Idea & R@1 & R@10 & R@20 & S@1 & S@10 & S@20 & R@1 & R@10 & R@20 & S@1 & S@10 & S@20 & R@1 & R@10 & R@20 & S@1 & S@10 & S@20 & R@1 & R@10 & R@20 & S@1 & S@10 & S@20 \\
            \hline
            ConcatEnc & 0.423 & 0.667 & 0.705 & 0.423 & 0.667 & 0.705 & 0.428 & 0.771 & 0.795 & 0.494 & 0.771 & 0.795 & 0.380 & 0.745 & 0.788 & 0.437 & 0.774 & 0.810 & 0.676 & 0.864 & 0.891 & 0.697 & 0.867 & 0.894 \\
            MultiSim & 0.282 & 0.654 & 0.705 & 0.282 & 0.654 & 0.705 & 0.367 & 0.693 & 0.783 & 0.410 & 0.723 & 0.807 & 0.263 & 0.622 & 0.679 & 0.305 & 0.646 & 0.703 & 0.497 & 0.812 & 0.840 & 0.521 & 0.814 & 0.840 \\
            \hline
            & \multicolumn{6}{c|}{Malay (msa)} & \multicolumn{6}{c|}{Portuguese (por)} & \multicolumn{6}{c|}{Spanish (spa)} & \multicolumn{6}{c}{Thai (tha)} \\
            Model & R@1 & R@10 & R@20 & S@1 & S@10 & S@20 & R@1 & R@10 & R@20 & S@1 & S@10 & S@20 & R@1 & R@10 & R@20 & S@1 & S@10 & S@20 & R@1 & R@10 & R@20 & S@1 & S@10 & S@20 \\
            \hline
            ConcatEnc & 0.538 & 0.871 & 0.924 & 0.571 & 0.876 & 0.924 & 0.423 & 0.823 & 0.863 & 0.500 & 0.831 & 0.871 & 0.538 & 0.860 & 0.898 & 0.569 & 0.865 & 0.902 & 0.619 & 0.952 & 0.976 & 0.619 & 0.952 & 0.976 \\
            MultiSim & 0.419 & 0.873 & 0.900 & 0.448 & 0.886 & 0.905 & 0.341 & 0.745 & 0.787 & 0.404 & 0.768 & 0.801 & 0.461 & 0.803 & 0.855 & 0.496 & 0.811 & 0.863 & 0.381 & 0.762 & 0.810 & 0.381 & 0.762 & 0.810 \\
            \hline
        \end{tabular}%
    }
    
    \vspace{1em}
    \resizebox{\textwidth}{!}{%
    \begin{tabular}{c|cccccc|cccccc|cccccc|cccccc}
        \multicolumn{25}{c}{Stage 3: Final System Evaluation Results} \\
        \hline
        & \multicolumn{6}{c|}{Arabic (ara)} & \multicolumn{6}{c|}{German (deu)} & \multicolumn{6}{c|}{English (eng)} & \multicolumn{6}{c}{French (fra)} \\
        Model & R@1 & R@10 & R@20 & S@1 & S@10 & S@20 & R@1 & R@10 & R@20 & S@1 & S@10 & S@20 & R@1 & R@10 & R@20 & S@1 & S@10 & S@20 & R@1 & R@10 & R@20 & S@1 & S@10 & S@20 \\
        \hline
        TriAligner & 0.359 & 0.705 & 0.731 & 0.359 & 0.705 & 0.731 & 0.464 & 0.795 & 0.873 & 0.542 & 0.795 & 0.880 & 0.374 & 0.769 & 0.819 & 0.418 & 0.793 & 0.835 & 0.676 & 0.856 & 0.894 & 0.697 & 0.856 & 0.894 \\
        TriAligner + Data Augmentation & 0.397 & 0.769 & 0.821 & 0.397 & 0.769 & 0.821 & 0.428 & 0.771 & 0.801 & 0.494 & 0.795 & 0.819 & 0.379 & 0.791 & 0.830 & 0.431 & 0.814 & 0.847 & 0.689 & 0.894 & 0.894 & 0.713 & 0.894 & 0.894 \\
        TriAligner + Augmentation + Re-Ranker & 0.481 & 0.822 & 0.821 & 0.481 & 0.822 & 0.821 & 0.468 & 0.778 & 0.801 & 0.556 & 0.797 & 0.819 & 0.392 & 0.804 & 0.830 & 0.448 & 0.822 & 0.847 & 0.781 & 0.897 & 0.894 & 0.805 & 0.897 & 0.894 \\
        \hline
        & \multicolumn{6}{c|}{Malay (msa)} & \multicolumn{6}{c|}{Portuguese (por)} & \multicolumn{6}{c|}{Spanish (spa)} & \multicolumn{6}{c}{Thai (tha)} \\
        Model & R@1 & R@10 & R@20 & S@1 & S@10 & S@20 & R@1 & R@10 & R@20 & S@1 & S@10 & S@20 & R@1 & R@10 & R@20 & S@1 & S@10 & S@20 & R@1 & R@10 & R@20 & S@1 & S@10 & S@20 \\
        \hline
        TriAligner & 0.556 & 0.890 & 0.933 & 0.590 & 0.895 & 0.933 & 0.433 & 0.867 & 0.897 & 0.510 & 0.877 & 0.907 & 0.589 & 0.875 & 0.914 & 0.629 & 0.883 & 0.919 & 0.571 & 0.952 & 0.976 & 0.571 & 0.952 & 0.976 \\
        TriAligner + Data Augmentation & 0.467 & 0.938 & 0.948 & 0.486 & 0.943 & 0.952 & 0.422 & 0.879 & 0.902 & 0.487 & 0.894 & 0.914 & 0.589 & 0.897 & 0.920 & 0.628 & 0.902 & 0.924 & 0.810 & 0.952 & 0.952 & 0.810 & 0.952 & 0.952 \\
        TriAligner + Augmentation + Re-Ranker & 0.436 & 0.923 & 0.948 & 0.457 & 0.933 & 0.952 & 0.425 & 0.896 & 0.902 & 0.486 & 0.914 & 0.914 & 0.652 & 0.920 & 0.920 & 0.694 & 0.924 & 0.924 & 0.904 & 0.952 & 0.952 & 0.904 & 0.952 & 0.952 \\

        \hline
    \end{tabular}%
    }

    \label{table_monolingual}
\end{table*}

\begin{table*}[t!]
    \centering
    \caption{Evaluation results at different stages for crosslingual settings, covering the eight most frequently used languages in the dataset.}
    \resizebox{\textwidth}{!}{%
        \begin{tabular}{c|cccccc|cccccc|cccccc|cccccc}
            \multicolumn{25}{c}{Stage 1: Baseline Evaluation Results} \\
            \hline
            & \multicolumn{6}{c|}{Arabic (ara)} & \multicolumn{6}{c|}{German (deu)} & \multicolumn{6}{c|}{English (eng)} & \multicolumn{6}{c}{French (fra)} \\
            Model/Source of Data & R@1 & R@10 & R@20 & S@1 & S@10 & S@20 & R@1 & R@10 & R@20 & S@1 & S@10 & S@20 & R@1 & R@10 & R@20 & S@1 & S@10 & S@20 & R@1 & R@10 & R@20 & S@1 & S@10 & S@20 \\
            \hline
            LaBSE/Native & 0.423 & 0.654 & 0.731 & 0.423 & 0.654 & 0.731 & 0.181 & 0.361 & 0.422 & 0.229 & 0.398 & 0.458 & 0.162 & 0.431 & 0.474 & 0.192 & 0.460 & 0.504 & 0.408 & 0.672 & 0.725 & 0.426 & 0.681 & 0.729 \\
            LaBSE/English & 0.500 & 0.667 & 0.833 & 0.500 & 0.667 & 0.833 & 0.000 & 0.350 & 0.550 & 0.000 & 0.400 & 0.600 & 0.120 & 0.426 & 0.462 & 0.139 & 0.481 & 0.506 & 0.140 & 0.580 & 0.680 & 0.160 & 0.600 & 0.680 \\
            BGE-M3/English & 0.500 & 0.833 & 0.833 & 0.500 & 0.833 & 0.833 & 0.150 & 0.550 & 0.550 & 0.200 & 0.600 & 0.600 & 0.190 & 0.559 & 0.620 & 0.203 & 0.582 & 0.646 & 0.300 & 0.500 & 0.560 & 0.320 & 0.520 & 0.560 \\
            BGE-M3/Native & 0.500 & 1.000 & 1.000 & 0.500 & 1.000 & 1.000 & 0.250 & 0.450 & 0.550 & 0.300 & 0.500 & 0.600 & 0.184 & 0.614 & 0.680 & 0.203 & 0.658 & 0.709 & 0.300 & 0.500 & 0.520 & 0.320 & 0.520 & 0.520 \\
            BGE-M3/Data Augmentation & 0.583 & 0.750 & 0.917 & 0.583 & 0.750 & 0.917 & 0.000 & 0.400 & 0.500 & 0.000 & 0.400 & 0.500 & 0.190 & 0.628 & 0.699 & 0.203 & 0.671 & 0.734 & 0.260 & 0.520 & 0.580 & 0.280 & 0.520 & 0.600 \\
            \hline
            & \multicolumn{6}{c|}{Malay (msa)} & \multicolumn{6}{c|}{Portuguese (por)} & \multicolumn{6}{c|}{Spanish (spa)} & \multicolumn{6}{c}{Thai (tha)} \\
            Model/Source of Data & R@1 & R@10 & R@20 & S@1 & S@10 & S@20 & R@1 & R@10 & R@20 & S@1 & S@10 & S@20 & R@1 & R@10 & R@20 & S@1 & S@10 & S@20 & R@1 & R@10 & R@20 & S@1 & S@10 & S@20 \\
            \hline
            LaBSE/Native & 0.157 & 0.600 & 0.667 & 0.171 & 0.619 & 0.676 & 0.187 & 0.589 & 0.643 & 0.222 & 0.629 & 0.675 & 0.322 & 0.575 & 0.637 & 0.340 & 0.590 & 0.654 & 0.548 & 0.881 & 0.905 & 0.548 & 0.881 & 0.905 \\
            LaBSE/English & 0.000 & 0.262 & 0.357 & 0.000 & 0.286 & 0.381 & 0.127 & 0.453 & 0.525 & 0.152 & 0.500 & 0.565 & 0.131 & 0.444 & 0.544 & 0.150 & 0.475 & 0.575 & 0.286 & 0.286 & 0.429 & 0.429 & 0.429 & 0.429 \\
            BGE-M3/English & 0.190 & 0.429 & 0.524 & 0.286 & 0.429 & 0.524 & 0.130 & 0.446 & 0.572 & 0.152 & 0.500 & 0.587 & 0.200 & 0.575 & 0.658 & 0.238 & 0.613 & 0.688 & 0.143 & 0.357 & 0.571 & 0.286 & 0.429 & 0.571 \\
            BGE-M3/Native & 0.143 & 0.619 & 0.667 & 0.190 & 0.619 & 0.667 & 0.185 & 0.703 & 0.754 & 0.196 & 0.739 & 0.783 & 0.263 & 0.644 & 0.675 & 0.300 & 0.675 & 0.700 & 0.071 & 0.071 & 0.214 & 0.143 & 0.143 & 0.286 \\
            BGE-M3/Data Augmentation & 0.214 & 0.476 & 0.571 & 0.286 & 0.476 & 0.571 & 0.123 & 0.616 & 0.699 & 0.174 & 0.652 & 0.717 & 0.269 & 0.513 & 0.613 & 0.313 & 0.563 & 0.638 & 0.071 & 0.571 & 0.571 & 0.143 & 0.571 & 0.571 \\
            \hline
        \end{tabular}%
    }

    \vspace{1em}
    \resizebox{\textwidth}{!}{%
    \begin{tabular}{c|cccccc|cccccc|cccccc|cccccc}
        \multicolumn{25}{c}{Stage 2: Ablation Study Results} \\
        \hline
        & \multicolumn{6}{c|}{Arabic (ara)} & \multicolumn{6}{c|}{German (deu)} & \multicolumn{6}{c|}{English (eng)} & \multicolumn{6}{c}{French (fra)} \\
        Idea & R@1 & R@10 & R@20 & S@1 & S@10 & S@20 & R@1 & R@10 & R@20 & S@1 & S@10 & S@20 & R@1 & R@10 & R@20 & S@1 & S@10 & S@20 & R@1 & R@10 & R@20 & S@1 & S@10 & S@20 \\
        \hline
        ConcatEnc & 0.333 & 0.667 & 0.667 & 0.333 & 0.667 & 0.667 & 0.250 & 0.900 & 0.900 & 0.300 & 0.900 & 0.900 & 0.327 & 0.786 & 0.827 & 0.354 & 0.810 & 0.848 & 0.380 & 0.660 & 0.680 & 0.400 & 0.680 & 0.680 \\
        MultiSim & 0.333 & 0.583 & 0.583 & 0.333 & 0.583 & 0.583 & 0.100 & 0.550 & 0.700 & 0.100 & 0.600 & 0.700 & 0.369 & 0.752 & 0.791 & 0.430 & 0.772 & 0.797 & 0.240 & 0.600 & 0.600 & 0.280 & 0.600 & 0.600 \\
        \hline
        & \multicolumn{6}{c|}{Malay (msa)} & \multicolumn{6}{c|}{Portuguese (por)} & \multicolumn{6}{c|}{Spanish (spa)} & \multicolumn{6}{c}{Thai (tha)} \\
        Model & R@1 & R@10 & R@20 & S@1 & S@10 & S@20 & R@1 & R@10 & R@20 & S@1 & S@10 & S@20 & R@1 & R@10 & R@20 & S@1 & S@10 & S@20 & R@1 & R@10 & R@20 & S@1 & S@10 & S@20 \\
        \hline
        ConcatEnc & 0.286 & 0.738 & 0.786 & 0.381 & 0.762 & 0.810 & 0.275 & 0.717 & 0.764 & 0.326 & 0.761 & 0.783 & 0.538 & 0.848 & 0.877 & 0.600 & 0.888 & 0.888 & 0.071 & 0.286 & 0.286 & 0.143 & 0.286 & 0.286 \\
        MultiSim & 0.476 & 0.881 & 0.881 & 0.571 & 0.905 & 0.905 & 0.156 & 0.732 & 0.808 & 0.196 & 0.761 & 0.826 & 0.381 & 0.785 & 0.846 & 0.425 & 0.813 & 0.863 & 0.000 & 0.143 & 0.286 & 0.000 & 0.286 & 0.571 \\
        \hline
    \end{tabular}%
}

    \vspace{1em}
    \resizebox{\textwidth}{!}{%
    \begin{tabular}{c|cccccc|cccccc|cccccc|cccccc}
        \multicolumn{25}{c}{Stage 3: Final System Evaluation Results} \\
        \hline
        & \multicolumn{6}{c|}{Arabic (ara)} & \multicolumn{6}{c|}{German (deu)} & \multicolumn{6}{c|}{English (eng)} & \multicolumn{6}{c}{French (fra)} \\
        Model & R@1 & R@10 & R@20 & S@1 & S@10 & S@20 & R@1 & R@10 & R@20 & S@1 & S@10 & S@20 & R@1 & R@10 & R@20 & S@1 & S@10 & S@20 & R@1 & R@10 & R@20 & S@1 & S@10 & S@20 \\
        \hline
        TriAligner & 0.333 & 0.667 & 0.667 & 0.333 & 0.667 & 0.667 & 0.250 & 0.900 & 0.900 & 0.300 & 0.900 & 0.900 & 0.327 & 0.786 & 0.827 & 0.354 & 0.810 & 0.848 & 0.380 & 0.660 & 0.680 & 0.400 & 0.680 & 0.680 \\
        TriAligner + Data Augmentation & 0.417 & 0.667 & 0.667 & 0.417 & 0.667 & 0.667 & 0.100 & 0.700 & 0.950 & 0.100 & 0.700 & 1.000 & 0.392 & 0.752 & 0.812 & 0.430 & 0.797 & 0.835 & 0.360 & 0.700 & 0.760 & 0.400 & 0.720 & 0.760 \\
        TriAligner + Augmentation + Re-Ranker & 0.417 & 0.667 & 0.667 & 0.417 & 0.667 & 0.667 & 0.150 & 0.950 & 0.950 & 0.200 & 1.000 & 1.000 & 0.367 & 0.780 & 0.812 & 0.417 & 0.810 & 0.835 & 0.440 & 0.760 & 0.760 & 0.480 & 0.760 & 0.760 \\
        \hline
        & \multicolumn{6}{c|}{Malay (msa)} & \multicolumn{6}{c|}{Portuguese (por)} & \multicolumn{6}{c|}{Spanish (spa)} & \multicolumn{6}{c}{Thai (tha)} \\
        Model & R@1 & R@10 & R@20 & S@1 & S@10 & S@20 & R@1 & R@10 & R@20 & S@1 & S@10 & S@20 & R@1 & R@10 & R@20 & S@1 & S@10 & S@20 & R@1 & R@10 & R@20 & S@1 & S@10 & S@20 \\
        \hline
        TriAligner & 0.286 & 0.738 & 0.786 & 0.381 & 0.762 & 0.810 & 0.275 & 0.717 & 0.764 & 0.326 & 0.761 & 0.783 & 0.538 & 0.848 & 0.877 & 0.600 & 0.888 & 0.888 & 0.071 & 0.286 & 0.286 & 0.143 & 0.286 & 0.286 \\
        TriAligner + Data Augmentation & 0.476 & 0.810 & 0.905 & 0.571 & 0.810 & 0.905 & 0.214 & 0.674 & 0.728 & 0.239 & 0.717 & 0.761 & 0.425 & 0.827 & 0.865 & 0.475 & 0.850 & 0.888 & 0.000 & 0.357 & 0.357 & 0.000 & 0.429 & 0.429 \\
        TriAligner + Augmentation + Re-Ranker & 0.333 & 0.857 & 0.905 & 0.380 & 0.857 & 0.905 & 0.217 & 0.684 & 0.728 & 0.282 & 0.717 & 0.761 & 0.475 & 0.827 & 0.865 & 0.525 & 0.850 & 0.888 & 0.071 & 0.285 & 0.357 & 0.142 & 0.428 & 0.429 \\
        \hline
    \end{tabular}%
}

    \label{table_crosslingual}
\end{table*}

\begin{table*}[t!]
    \centering
    % Revised Caption: Mentioning S@10 explicitly
    \caption{Comparison of S@10 scores with the first-place team on the competition's test set for the monolingual task.}
    \label{tab:winner_comparison_mono} % Kept the label for referencing
    \resizebox{\textwidth}{!}{% Use resizebox to ensure it fits within the text width
        \begin{tabular}{l|ccccccccccc} % 'l' for left-aligned team name, 'c' for centered metrics
            \hline
            % Revised Headers: Language (code) format
            Team Name & Average & English (eng) & French (fra) & German (deu) & Portuguese (por) & Spanish (spa) & Thai (tha) & Malay (msa) & Arabic (ara) & Turkish (tur) & Polish (pol) \\
            \hline
            First place team & \textbf{0.960} & 0.916 & 0.972 & 0.958 & 0.926 & 0.974 & 0.994 & 1.000 & 0.986 & 0.948 & 0.926 \\
            MultiMind & \textbf{0.808} & 0.674 & 0.864 & 0.800 & 0.748 & 0.776 & 0.923 & 0.957 & 0.848 & 0.746 & 0.744 \\
            \hline
        \end{tabular}%
    }
\end{table*}

\end{document}